\documentclass[sigconf]{acmart}

\AtBeginDocument{%
  }

\setcopyright{none}
\renewcommand\footnotetextcopyrightpermission[1]{}

\thanks{This is an extended version of the UMAP '26 Industry Track paper: \url{https://doi.org/10.1145/3774935.3807910}. Copyright held by the author(s).}%

\acmConference[UMAP '26]{34th ACM International Conference on User Modeling, Adaptation and Personalization}{June 8--11, 2026}{Gothenburg, Sweden}

\usepackage{bm}
\usepackage{enumitem}
\usepackage{multirow}

\DeclareMathOperator{\PR}{\mathbb{P}}

\newcommand{\R}{\mathbb{R}}
\newcommand{\att}{\mathrm{A \scriptstyle TT}}

\begin{document}

\title{Long-Term Embeddings for Balanced Personalization}

\author{Andrii Dzhoha}
\email{andrew.dzhoha@zalando.de}
\affiliation{%
  \institution{Zalando SE}
  \city{Berlin}
  \country{Germany}
}

\author{Egor Malykh}
\email{egor.malykh@zalando.de}
\affiliation{%
  \institution{Zalando SE}
  \city{Berlin}
  \country{Germany}
}

\begin{abstract}
  Modern transformer-based sequential recommenders excel at capturing short-term intent but often suffer from recency bias, overlooking stable long-term preferences. While extending sequence lengths is an intuitive fix, it is computationally inefficient, and recent interactions tend to dominate the model's attention. We propose Long-Term Embeddings (LTE) as a high-inertia contextual anchor to bridge this gap. We address a critical production challenge: the point-in-time consistency problem caused by infrastructure constraints, as feature stores typically host only a single ``live'' version of features. This leads to an offline-online mismatch during model deployments and rollbacks, as models are forced to process evolved representations they never saw during training. To resolve this, we introduce an LTE framework that constrains embeddings to a fixed semantic basis of content-based item representations, ensuring cross-version compatibility. Furthermore, we investigate integration strategies for causal language modeling, considering the data leakage issue that occurs when the LTE and the transformer's short-term sequence share a temporal horizon. We evaluate two representations: a heuristic average and an asymmetric autoencoder with a fixed decoder grounded in the semantic basis to enable behavioral fine-tuning while maintaining stability. Online A/B tests on Zalando demonstrate that integrating LTE as a contextual prefix token using a lagged window yields significant uplifts in both user engagement and financial metrics.
\end{abstract}



\maketitle

\section{Introduction}

Modeling sequential interactions is essential in modern applications such as e-commerce, music streaming, and video platforms, where past user behaviors help predict future recommendations. Historically, the industry approached this using structured or tabular data with models such as gradient-boosted decision trees and logistic regression. These relied on state-based features -- hand-crafted counters summarizing long-term behavior into continuous values, such as the number of items from a specific brand purchased in the past year. The field has since shifted toward sequence-based transformer architectures (e.g., SASRec~\cite{Kang2018SelfAttentiveSR}). These models often frame recommendation as a causal language modeling (CLM) task, excelling at capturing fluid, short-term user intent due to their efficiency and effectiveness~\cite{dzhoha2025efficienteffectivequerycontextaware, 10.1145/3336191.3371786, 10.5555/3367471.3367642}.

However, this shift has created a gap in long-term representation. While transformers handle continuous inputs, they are primarily optimized for discrete item sequences and face practical limits in large-scale settings. First, the $O(N^2)$ self-attention mechanism makes it computationally prohibitive to use a customer's entire multi-year history -- potentially thousands of interactions -- in real-time ranking. Second, by focusing on the most recent items, transformers lose the broader context of long-term behavior. This leads to recency bias~\cite{oh2024measuring, chang2022recency}: models are highly sensitive to immediate intent (e.g., searching for a red dress) but often forget stable preferences, such as a liking for premium brands or specific clothing sizes.

We propose long-term embeddings (LTEs) as a compressed memory or anchor for long-term affinity. While transformers handle ``what now'' (short-term intent), LTEs provide a stable signal for ``who'' the user is (long-term preference). By summarizing the distant past into a single vector, LTEs bypass the quadratic bottleneck of attention and keep a global user profile present, even when the user is casually browsing. Beyond performance, LTEs serve as a universal, downstream-agnostic signal. By capturing stable preferences, they can be integrated into various models -- from homepage personalization to newsletters -- without task-specific retraining.

Despite their systemic value, deploying LTEs at scale introduces infrastructure and operational challenges often overlooked in research. While it is feasible to store multiple versions of LTEs for offline training, online serving is fundamentally constrained. For a user base in the tens of millions, a single version of LTEs can reach terabytes of data. High-performance feature stores typically restrict the system to hosting only the latest snapshot of the embedding table for real-time use~\cite{pmlr-v67-li17a}. This ``single-version'' setup creates a significant point-in-time consistency problem, resulting in an offline-online mismatch:
\begin{itemize}
  \item \textbf{Training-serving skew}: Models are trained on historical logs (offline), but the feature store provides only ``live'' versions (online). If a ranker is trained on an LTE from day $T$ but deployed on day $T+2$, it receives a signal that has evolved beyond its training distribution.
  \item \textbf{Version mismatch during rollbacks}: During a production incident, rolling back to an older model is a standard mitigation. However, if the feature store has already updated to the latest LTEs, the old model must process representations it never saw during training. Rewinding terabyte-scale tables in real-time is often technically impossible, forcing a choice between a broken new model or an uncalibrated old one.
\end{itemize}

We address these challenges by introducing a \textit{high-inertia LTE framework} that constrains long-term embeddings to a \textit{fixed semantic basis}. By representing each LTE as a linear combination of static content-based embeddings, we ensure that the latent space remains stable and compatible across both time and model versions. This framework employs a lagged sliding window, delaying the history used for LTEs relative to the recent sequence processed by the transformer. This design prevents data leakage in CLM modeling and provides a stability buffer for production. Specifically, we investigate:
\begin{itemize}
  \item \textbf{Representation}: Methods for obtaining LTEs that satisfy feature store constraints, including (i) a heuristic average of content-based embeddings and (ii) an asymmetric autoencoder that learns to reconstruct user history by mapping behavioral data into a fixed content-embedding space.
  \item \textbf{Integration}: Strategies for integrating these anchors into CLM-based ranking models to effectively balance short-term intent and long-term preference.
\end{itemize}

Our contributions include:
\begin{enumerate}
  \item \textbf{High-inertia framework}: We propose a long-term embedding (LTE) framework that utilizes a lagged window and a fixed semantic basis to solve the point-in-time consistency problem in production environments.
  \item \textbf{Integration strategies}: We investigate multiple fusion methods for integrating LTE into causal language models, identifying contextual anchoring as the superior strategy for balancing short-term intent and long-term preference.
  \item \textbf{Attention migration analysis}: We provide a granular empirical analysis of how LTE redistributes the transformer's attention budget, demonstrating a significant reduction in recency bias and the reclamation of distant historical memory.
  \item \textbf{Behavioral fine-tuning}: We introduce an asymmetric autoencoder with a fixed decoder that enables learning behavioral affinities (e.g., price-range, style) while strictly maintaining the semantic stability required for model rollbacks.
  \item \textbf{Large-scale validation}: We validate our approach through extensive offline experiments and online A/B tests on a platform with millions of users, demonstrating significant uplifts in both engagement and financial metrics.
\end{enumerate}

\section{Related work}

The challenge of balancing long-term user profiles with short-term intent has been a focal point of recent research. We categorize the literature into three main areas: sequential modeling, lifelong user representation, and the industrial constraints of embedding stabilization.

\subsection{Sequential and transformer-based recommendation}

The transformer architecture~\cite{10.5555/3295222.3295349} revolutionized sequential recommendation by addressing long-range dependencies more effectively than previous recurrent approaches. State-of-the-art models like SASRec~\cite{Kang2018SelfAttentiveSR} and BERT4Rec~\cite{10.1145/3357384.3357895} utilize self-attention to capture dependencies within interaction sequences, primarily focusing on next-item prediction~\cite{10.1145/3190616}. However, these models are typically constrained to short sequences due to the quadratic complexity of attention. As we demonstrate in Section~\ref{sec:extended_sequences}, naively extending these sequences introduces significant computational overhead with marginal gains, as dominant recent interactions mask distant stylistic signals. Our work builds on these architectures by introducing the LTE as a contextual anchor, preserving global preferences without increasing the input sequence length or the online serving latency associated with larger payloads and feature store lookups.

\subsection{Lifelong and long-term user modeling}

To capture extended histories, researchers have proposed memory-augmented networks like MIMN~\cite{10.1145/3292500.3330666} and HPMN~\cite{10.1145/3331184.3331230}, which maintain external states to track evolving interests. Others, like PinnerSage~\cite{10.1145/3394486.3403280}, employ clustering to represent multi-faceted interests over long horizons. Recent industrial efforts like TransActV2~\cite{10.1145/3746252.3761433} manage multi-year sequences but rely on top-$k$ retrieval steps per sample, shifting complexity to high-latency retrieval. Similarly, DMT~\cite{10.1145/3340531.3412697} maintains separate encoders for different behavior types, increasing the parameter footprint. While models like LONGER~\cite{10.1145/3705328.3748065} and GPSD~\cite{10.1145/3711896.3737117} focus on the learning dynamics of extended sequences via generative pre-training, they require massive resources for a monolithic sequence. In contrast, our approach avoids $O(N^2)$ scaling and per-sample similarity searches by utilizing a content-grounded LTE that remains lightweight for real-time ranking.

\subsection{Industrial constraints and embedding stabilization}

A significant gap exists between academic modeling and the operational reality of serving features at scale. Industrial deployments face a triad of constraints: single-versioned feature stores, strict latency budgets, and the need for point-in-time consistency during model rollbacks~\cite{pmlr-v67-li17a}. Recent research has addressed embedding instability across retraining cycles via post-hoc transformations. For instance, \cite{10.1145/3705328.3748141} utilize Orthogonal Procrustes and SVD to map new embeddings into a standardized reference space.

While such methods are mathematically sound, they introduce significant industrial overhead: they require maintaining a seed training run as a permanent dependency, increase operational complexity by requiring the storage and retrieval of transformation matrices for every version, and struggle with item turnover in cold-start scenarios. Our work differs by treating stability as a primary design goal rather than a post-processing task. By grounding LTEs in a \textit{fixed semantic basis} -- leveraging content-based foundation models like CLIP~\cite{clip-2021} -- we achieve what we term \textit{temporal inertia}. This framework aims to improve model robustness and cross-version compatibility without adding complexity to downstream models, nor increasing inference or training latency, and does so without the need for external historical referencing or complex transformation pipelines.

\section{Methodology}

\subsection{Problem statement}

We address the sequential recommendation problem: predicting the next item a user will interact with, given their historical sequence of interactions. Formally, for user $u$ with interaction history $\left(x_i^{u}\right)_{i=1}^{N+1}$, and a long-term user signal $l_{i+1}^{u}$, the goal is to estimate the probability distribution over candidate items $x \in \mathcal{X}$:
\begin{equation*}
  \PR\left(x_{i+1}^{u} = x \;\middle|\; x_1^{u},\dots,x_i^{u};\,l_{i+1}^{u}\right).
\end{equation*}
Here, $N$ is the number of observed interactions. The $(N+1)$-th item serves as the prediction target.

\subsection{Method}\label{sec:method}

We employ a deep self-attention transformer model trained with causal language modeling, adopting the SASRec-style architecture~\cite{Kang2018SelfAttentiveSR, 10.1145/3336191.3371786, 10.5555/3367471.3367642}. The model comprises an item embedding layer, multiple self-attention blocks (featuring multi-head attention, feed-forward layers, residual connections, and layer normalization~\cite{ba2016layer, nguyen2019transformers, 10.5555/3295222.3295349}), and an output projection.

Given input $\bm{X}^{(h)} \in \R^{N \times D}$ at layer block $h$, the output is:
\begin{equation}\label{eq:attention}
    \bm{X}^{(h+1)} = \att\left(\bm{X}^{(h)}\right),
\end{equation}
where $\att$ refers to the attention layer block operations described above, with attention restricted to previous positions by causal masking.

The model efficiently computes relevance scores for all candidate items at each sequence position in a single forward pass through $H$ self-attention blocks:
\begin{equation}\label{eq:scores-matrix}
  \bm{Y}^u = \bm{X}^{(H+1)} \bm{W}_O \bm{A}^{\intercal} \in \R^{N \times |\mathcal{X}|},
\end{equation}
where $\bm{W}_O$ is the output projection and $\bm{A} \in \R^{|\mathcal{X}| \times D}$ is the item embedding matrix.

Training uses categorical cross-entropy loss~\cite{di2024theoretical}. To handle variable-length sequences, left padding is applied and masked during both attention and loss computation. At inference, the model uses the final sequence position's representation to predict the next item given the observed history $\left(x_i^u\right)_{i=1}^{N}$, or to rank a candidate set via dot-product relevance scores.

\subsection{Long-term user signals}

To incorporate long-term user preferences, we explore several integration strategies for the long-term embedding $l_{i+1}^{u}$:
\begin{enumerate}
  \item Combining outside the transformer, at the output projection stage.
  \item Prepending as a contextual prefix token to the input sequence.
  \item Adding to each item embedding in the sequence.
\end{enumerate}
These approaches allow the model to attend to both recent interactions and stable user preferences.

To address point-in-time consistency and rollback challenges in production, we constrain each long-term embedding $l_{i}^{u}$ to a \textit{fixed semantic space}. Specifically, it is computed as a linear combination of static content-based embeddings. This design ensures that LTEs remain stable and compatible across time and model versions, providing robust representations for online serving.

\section{Approach}
\label{sec:approach}

We argue that representing long-term embeddings via a weighted average of content-based item embeddings yields a meaningful and expressive feature for sequential recommendation. These content-based embeddings (e.g., CLIP-based representations~\cite{clip-2021}) are derived from intrinsic item attributes such as category, brand, and style, capturing properties independent of user behavior. In our preliminary exploratory analysis, we found that such representations exhibit significant expressiveness; style mapping and semantic variance analysis showed that users with similar high-level behaviors cluster together naturally in this space.

We consider two primary weighting schemes for the LTE calculation: a uniform average and a recency-weighted average, where recent interactions are given higher weights. Using these representations, we study three architectures to integrate these signals into the CLM-based transformer ranker.

\subsection{Late fusion: LTE outside the transformer}

In this approach, the LTE is computed over the $[365, 0]$ window, meaning using the last 365 days of interactions, and integrated at the output projection stage by concatenating it with the final transformer output at all sequence positions before the dot-product with item embeddings. Modifying (\ref{eq:scores-matrix}), the relevance score matrix is computed as:
\begin{equation*}
  \bm{Y}^u = (\bm{X}^{(H+1)} + \bm{1}_N \bm{L}_u) \bm{W}_O \bm{A}^{\intercal},
\end{equation*}
where $\bm{1}_N \in \mathbb{R}^{N \times 1}$ is an all-ones vector. A limitation here is effectively bypassing the self-attention layers, preventing the model from learning complex dependencies between the long-term profile and short-term sequence.

\subsection{Contextual anchoring: LTE as a prefix token}

In this method, the LTE acts as a global context token. By prepending it, the sequence length increases to $N+1$. To maintain the causal property, this token is placed at index $0$ so that all subsequent items $x_i$ can attend to it. The augmented input matrix for the first attention block (\ref{eq:attention}) is defined as:
\begin{equation*}
  \bm{X}^{(2)} = \att\left(\left[\bm{L}_u ; \bm{X}^{(1)}\right]\right),
\end{equation*}
where $[;]$ denotes row-wise concatenation. This ensures that item representations at every layer are conditioned on the long-term user profile.

In a causal language modeling setup, the LTE token at position $0$ influences all subsequent hidden states through the self-attention mechanism. Consequently, computing the LTE on the same temporal horizon as the short-term sequence would introduce data leakage. To prevent this, we compute the LTE over a lagged window $[365, T]$, excluding the most recent $T$ days of interactions (e.g., 60 days) used in the transformer's input sequence.

\subsection{Feature injection: LTE added to item embeddings}

Here, the LTE is added to each item embedding in the sequence (\ref{eq:attention}):
\begin{equation*}
  \bm{X}^{(2)} = \att\left(\bm{X}^{(1)} + \bm{1}_N \bm{L}_u\right).
\end{equation*}

This approach injects the long-term signal not only into the value position (as in the prefix token method) but also into the query and key positions of the attention mechanism. This enables the model to modulate attention scores based on both the current item and the user's preferences, conditioning item-to-item attention on long-term information. As with the prefix token method, a lagged $[365, T]$ window is required to prevent data leakage.

\section{Offline experiments}

We evaluate the proposed LTE integration methods and weighting schemes using a ranking model that powers Browse and Search for Zalando across 25 markets.

\subsection{Base model}

Our baseline adopts a two-tower architecture~\cite{10.1145/2959100.2959190, dzhoha2024reducingpopularityinfluenceaddressing}, following the training procedure outlined in Section~\ref{sec:method}. The user tower is a transformer with two residual blocks and four-head multi-head attention ($H=2$, $d_{head}=4$) while the item tower is a feed-forward network producing the item embeddings used in $\bm{A}$ from (\ref{eq:scores-matrix}). The towers are trained jointly but deployed independently: item embeddings are indexed in a vector store for retrieval, and the user tower generates real-time user embeddings from the last 100 interactions (within the 60-day window) for nearest-neighbor search. Training uses sampled softmax loss with log-uniform sampling (0.5\% negative classes). Each item input is formed by concatenating the embeddings of the interacted item, interaction type, and categorical timestamp.

\subsection{Dataset and evaluation protocol}

Our short-term sequence dataset consists of item interactions from the past 60 days, including clicks, add-to-wishlist, add-to-cart, and checkout events, all attributed to Catalog Browse and Search scenarios using a last-touch attribution model. Each interaction is joined with the corresponding item ID, timestamp, and interaction type, then sorted chronologically. The training set contains over 70 million unique users across 25 markets, and the evaluation set includes more than 1,000,000 users. To avoid data leakage, we enforce a strict temporal split, evaluating on a holdout set from the day immediately following the training period, mirroring our daily production retraining cycle. Performance is assessed using normalized discounted cumulative gain at cutoff 500 (NDCG@500), calculated for each next-item prediction in the holdout day to measure the quality of the ranked recommendations.

\subsection{Limitation of extended sequence lengths}
\label{sec:extended_sequences}

As a preliminary experiment, we increased the base transformer model's sequence length from 60 to 360 days, incrementally extending both the months of data and the sequence length. Marginal improvements in NDCG were observed after two months (sequence length 100), with gains stalling beyond that. Computational costs rose sharply: training time increased at least fivefold, and data preparation became eight times slower. We also found that effectively leveraging such long sequences would require substantially larger model capacity to handle the distributional shifts over extended periods. While ongoing research explores efficient transformer variants for long sequences, these approaches demand significant architectural changes and would increase inference latency. These findings motivated us to pursue long-term embeddings as a compressed memory or anchor for long-term user affinity, rather than simply extending sequence lengths.

\subsection{Integration results and discussion}

Here, we present the offline evaluation results comparing the proposed LTE integration methods from Section~\ref{sec:approach}. To ensure compatibility, each LTE is projected into the same space as the item embeddings using a linear layer. The LTE is computed over the specified window by averaging content-based embeddings of items the user interacted with during that period. As content-based embeddings, we use CLIP-based representations~\cite{clip-2021}: a multimodal vision-language model that maps product images and descriptions into a shared 512-dimensional semantic space. Our product images and descriptions are encoded and aggregated into a single embedding per item. For customers without sufficient long-term interactions, we use a zero vector as the default LTE. Across variants, we experimented with different projection architectures (one or two layers, linear or GELU activations, and intermediate dimensions of 128, 256, 512, or 1024), reporting the best-performing configuration. For the recency-weighted average, we applied exponential decay to emphasize recent interactions.

Table~\ref{tab:offline_results} summarizes the results. All gains over the baseline are statistically significant ($p < 0.05$) via paired t-test.

\begin{table}[ht]
\centering
\caption{NDCG@500 Uplift for LTE integration strategies.}
\label{tab:offline_results}
\begin{tabular}{@{}lccc@{}}
\toprule
\textbf{Integration method} & \textbf{Window} & \textbf{Uniform} & \textbf{Recency} \\ \midrule
Late fusion (outside)       & $[365, 0]$      & -0.70\%          & -3.05\%                   \\
Feature injection (added)    & $[365, 60]$     & +0.45\%          & +0.42\%                   \\
Contextual anchoring & $[365, 60]$ & +1.31\% & +0.87\%                   \\ \bottomrule
\end{tabular}
\end{table}

The results yield several key insights:
\begin{itemize}
  \item \textbf{Contextual anchoring is superior:} Integrating LTE as a prefix token with a lagged window $[365, 60]$ yields the highest NDCG@500 improvement (+1.31\% for the uniform average), confirming that treating the long-term signal as a global context allows the self-attention mechanism to effectively condition short-term intent on stable preferences, acting as a compressed memory of the pre short-term sequence.
  \item \textbf{Uniform vs. recency-weighted LTE:} Recency-weighted averages consistently underperform uniform averages across all integration strategies. One plausible explanation is that the transformer's sequence modeling already captures short-term recency, so emphasizing it in the LTE introduces redundancy, whereas a uniform average may provide a more complementary long-term signal. However, other factors -- such as the interaction between exponential decay rates and window length -- may also contribute, and further ablation is needed to fully disentangle the effect.
  \item \textbf{Limitations of late fusion and feature injection:} Integrating LTE outside the transformer leads to performance degradation, suggesting the importance of fusing long-term preferences within the self-attention layers to capture complex dependencies. Adding LTE to item embeddings (feature injection) also underperforms the prefix token approach, likely because injecting the long-term signal into query and key positions introduces noise that dilutes the model's focus on relevant short-term interactions.
\end{itemize}

\subsection{Data leakage ablation study}

To quantify the effect of data leakage when the LTE and short-term sequence share a temporal horizon in CLM modeling, we conducted an ablation study using the prefix token integration method. We compared two scenarios: (i) LTE computed over a lagged window $[365, 60]$ to prevent data leakage, and (ii) LTE computed over the full window $[365, 0]$, which includes recent interactions overlapping with the short-term sequence, allowing the model to attend to future information via the LTE token. While the prefix token approach still improved performance, using the full window resulted in a 0.5\% drop in NDCG compared to the lagged window, confirming that overlapping windows introduce data leakage and hinder generalization.

\section{Analysis of attention redistribution}
\label{sec:attention_analysis}

To understand how LTE alters the transformer's decision-making, we analyze the attention allocation of the final layer ($H=2$). We focus on how the model balances immediate intent against the broader historical context.

\subsection{Recency intensity}

Sequential recommenders often suffer from an over-reliance on the most recent interactions, creating a filter bubble of immediate intent. We quantify this through recency intensity: the ratio of attention key weight assigned to the last interaction ($x_N$) relative to an earlier interaction ($x_{N-49}$).

By analyzing a holdout set of users with history lengths of at least 50 items, we observe that the baseline model exhibits a recency intensity of $2.83$. Upon integrating the LTE, this ratio drops to $2.62$ -- a $7.64\%$ reduction in recency bias. This shift indicates that the LTE acts as a high-inertia anchor, allowing the model to de-prioritize potentially noisy immediate clicks in favor of a more balanced historical view.

\subsection{Attention migration}

To isolate how the model redistributes its attention budget, we analyze the relative change in focus across the interaction sequence, termed the attention migration delta. This analysis allows us to visualize the migration of energy within the model: identifying exactly which items lost attention so that others could gain it.

\begin{figure}[t]
  \centering
  \includegraphics[width=1.0\linewidth]{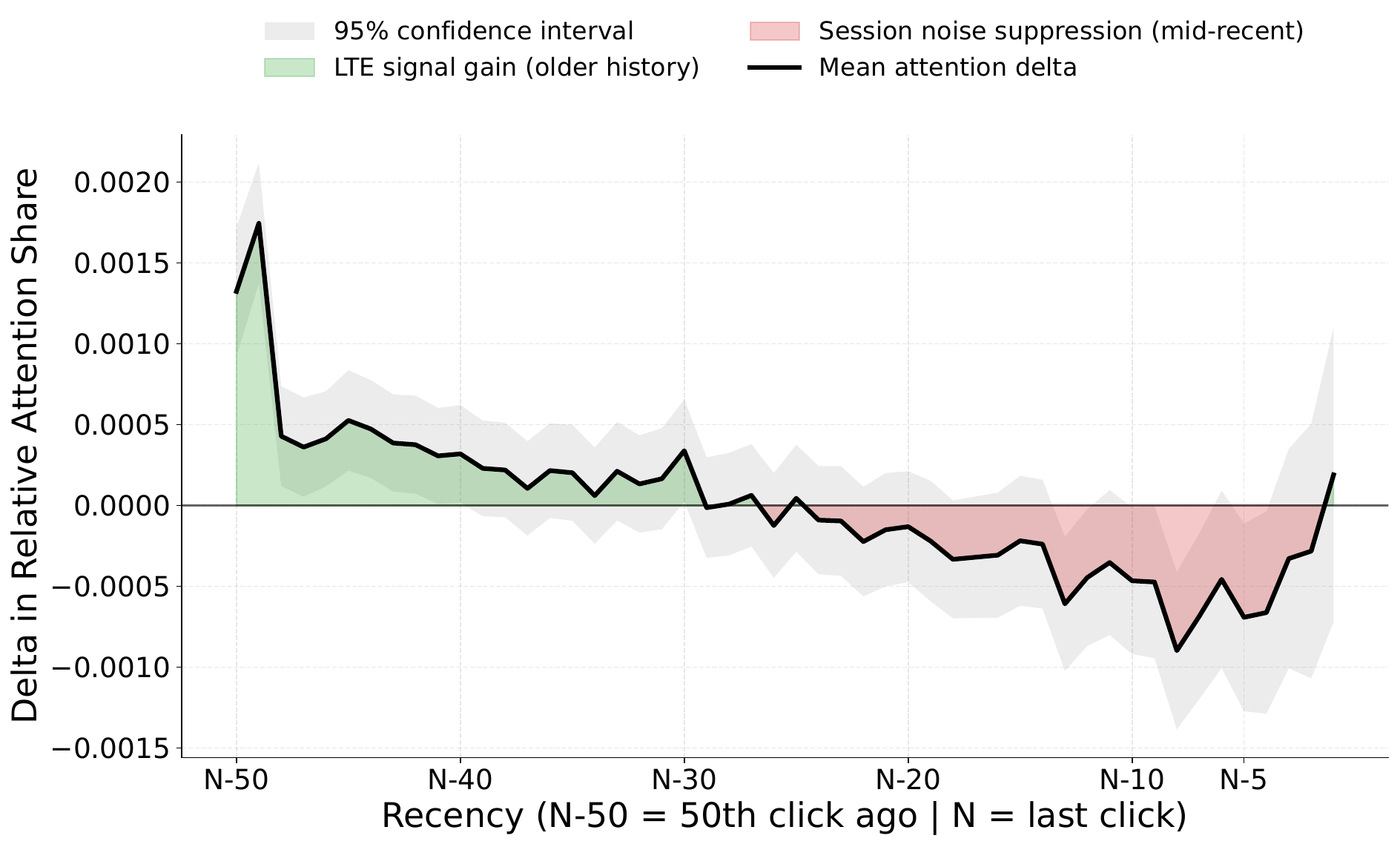}
  \Description{Bar chart showing the attention migration delta (LTE minus baseline) over the last 50 interaction positions. Older positions (x_{N-49} to x_{N-25}) show positive deltas indicating increased attention, while mid-recent positions (x_{N-24} to x_{N-1}) show negative deltas indicating decreased attention. Gray bands indicate 95\% confidence intervals.}
  \caption{Statistical consistency of LTE attention migration. The delta (LTE - baseline) is computed after normalizing attention weights within the most recent 50 interactions to a local share of 1.0. The model appears to moderate mid-session fluctuations to reclaim attention share for older history. Gray bands indicate the 95\% confidence interval calculated across the holdout set.}
  \label{fig:attention_delta}
\end{figure}

For every user, we represent the attention assigned to the last 50 interactions as a relative share of a fixed total. This normalization ensures that the comparison is not skewed by the total magnitude of attention, but rather reflects the internal priority shift of the transformer. By comparing the average attention share at each point in the sequence -- calculated with respect to the holdout set -- we can identify where the LTE model invests more energy and where it withdraws it relative to the baseline. As illustrated in Figure~\ref{fig:attention_delta}, two plausible patterns emerge:
\begin{itemize}
  \item \textbf{Memory reclamation}: The model appears to shift more attention to older interactions (positions $x_{N-49}$ to $x_{N-25}$), recovering them from the transformer's typical attention decay and allowing them to remain influential in the final prediction.
  \item \textbf{Noise suppression}: Conversely, the model reduces focus on the mid-recent interactions ($x_{N-24}$ to $x_{N-1}$), suggesting it downweights session-level fluctuations that do not align with the user's stable, long-term identity.
\end{itemize}

\section{Analysis of temporal stability and production resilience}
\label{sec:stability}

While the $[365, 60]$ lagged window prevents data leakage, it introduces a potential version mismatch during production rollbacks. This mismatch arises because both the LTE model and the downstream ranker typically follow a synchronized retraining schedule (e.g., daily). If a rollback reverts the ranker to a model artifact from $X$ days earlier, it must process ``future'' LTE features from the feature store that were not present during its original training. To quantify the impact of this version mismatch, we analyze it through the concept of \textit{temporal inertia}, which measures the stability of user profiles over time.

\subsection{Theoretical drift bound}

We define the turnover rate ($\tau$) as the fraction of the interaction set that changes within the user's history over a rollback period of $X$ days:
\begin{equation*}
  \tau = \frac{|S_{in}| + |S_{out}|}{N},
\end{equation*}
where $N$ is the total number of items in the 305-day window, $S_{in}$ is the set of items entering the window, and $S_{out}$ the set of items exiting. The drift between the version expected by the ranker ($\text{LTE}_{t-X}$) and the current production version ($\text{LTE}_t$) is bounded by:
\begin{equation*}
  \|\text{LTE}_t - \text{LTE}_{t-X}\| \le \tau \cdot \max \|e\|,
\end{equation*}
with $e$ representing an individual content-based item embedding. In the case of unit-normalized embeddings, the maximum drift is simply $\tau$.

Assuming a roughly uniform distribution of shopping activity over the year, the turnover for a short rollback window (e.g., $X \le 10$ days) is small. Furthermore, we argue that because user behavior patterns exhibit strong consistency, new interactions often align with established preferences. Consequently, even for low-activity users where $\tau$ is higher, the resulting vector remains semantically proximal to the previous version, preventing a drastic shift in the user's latent profile.

\subsection{Evaluation protocol and resilience}

To isolate the specific impact of LTE version mismatch, we must distinguish it from general model staleness (e.g., degradation due to shifting catalog trends). We compare the performance of a model using LTE against a baseline transformer that relies solely on short-term sequences.

We define relative resilience as the difference in NDCG degradation between the LTE-augmented model and the baseline. If the LTE model exhibits lower decay than the baseline, it indicates that the long-term signal acts as a stabilizing contextual anchor.

\begin{table}[ht]
\centering
\caption{Stability and resilience of LTE under version mismatch.}
\label{tab:stability_results}
\begin{tabular}{@{}cccc@{}}
\toprule
\textbf{Rollback} & \textbf{Avg. turnover} & \textbf{Mean} & \textbf{Relative} \\
\textbf{offset ($X$)} & \textbf{rate ($\tau$)} & \textbf{cosine sim.} & \textbf{resilience} \\ \midrule
1 Day             & 0.7\%                  & 0.997         & +0.69\%         \\
5 Days            & 2.8\%                  & 0.994         & +1.01\%         \\
10 Days           & 5.4\%                  & 0.985         & +1.65\%         \\ \bottomrule
\end{tabular}
\end{table}

As shown in Table~\ref{tab:stability_results}, the relative resilience is positive and increases with the rollback offset. While both models naturally degrade over time, the LTE-augmented model is significantly more robust. The high mean cosine similarity indicates that LTEs stay within the semantic manifold the transformer was trained to recognize. This ``stability-by-design'' allows us to maintain a single-versioned feature store without sacrificing performance during production incidents or deployments.

\section{Online experiment}

The promising offline results led to the deployment of the LTE framework in a large-scale online A/B test. We implemented the LTE as a prefix token using the lagged $[365, 60]$ window with a uniform average weighting scheme, as this configuration demonstrated the optimal balance of performance and temporal stability.

The experiment was conducted on the ranking systems for the Browse and Search use cases across 25 markets, involving millions of active users\footnote{Our data collection process complies with the regulations defined in the GDPR and other existing regulatory frameworks around data privacy and safety in the European Union.}. While Search optimizes for findability through explicit queries, Browse facilitates open-ended exploration via navigation and personalized feeds. Both use cases were served by a highly optimized production baseline prior to LTE integration. We utilized equal traffic splits to ensure sufficient power to detect a minimum detectable effect (MDE) for our primary KPIs, with statistical significance defined at $p < 0.05$.

The A/B test results are summarized in Table~\ref{tab:online_results}. We define engagement through high-value user actions (e.g., add-to-wishlist, add-to-cart) and revenue as the net merchandise volume per user after returns.

\begin{table}[ht]
\caption{A/B test results for integrating LTE into the ranking model. Engagement encompasses high-value actions (add-to-wishlist, add-to-cart). Revenue represents net merchandise volume per user.}
\label{tab:online_results}
\centering
\begin{tabular}{l c c c c}
\toprule
& \multicolumn{3}{c}{\textbf{Engagement}} & \\
\cmidrule(lr){2-4}
& Browse & Search & All & \textbf{Revenue} \\
\midrule
Estimate & +1.16\% & +0.15\% & +0.61\% & +0.42\% \\
95\% CI & [0.79, 1.53]\% & [-0.2, 0.5]\% & [0.32, 0.9]\% & [0.07, 0.76]\% \\
\bottomrule
\end{tabular}
\end{table}

The results indicate that the Browse experience was most positively impacted, showing a significant +1.16\% uplift in engagement. In contrast, the Search use case showed a marginal, non-significant improvement. This suggests that in Search, the explicit, fine-grained query intent is the primary driver of relevance, often superseding long-term stylistic preferences. However, in Browse -- where user intent is more latent and explorative -- the LTE acts as a critical anchor that aligns the recommendations with the user's historical style.

Overall, combined Browse and Search engagement improved by 0.61\%, accompanied by a +0.42\% increase in revenue. These results confirm the practical value of high-inertia LTEs in balancing short-term intent with long-term preferences in a high-traffic production environment.

\section{Behavioral fine-tuning of long-term embeddings}

In the initial deployment, we introduced a high-inertia LTE framework that integrates long-term user signals into a transformer-based ranker using a CLM procedure. The LTE vector, computed as a prefix token over a lagged one-year window to prevent data leakage, led to measurable uplifts in both engagement and financial metrics in online experiments. This signal was constructed by averaging CLIP-based content embeddings for each customer over a year of interactions. However, this simple averaging approach faces two main limitations:
\begin{enumerate}
  \item It treats all historical interactions with equal importance, failing to distinguish between fleeting clicks and interactions that define a user's core profile.
  \item Content-based signals alone often lack exposure to intrinsic customer affinities, such as price-point sensitivity or quality preferences, which can only be derived from behavioral data.
\end{enumerate}

To address these issues, we propose a fine-tuning approach that learns to weight and adjust content-based embeddings using behavioral data, while preserving the high-inertia properties necessary for production resilience. Rather than relying on transformer-based architectures -- which are inherently complex and primarily capture short- to mid-term intent -- we introduce an \textit{asymmetric autoencoder with a fixed semantic basis} to focus on long-term user preferences.

\subsection{Architecture and objective}

The model represents a user's history as a sparse multi-hot vector $\bm{h}_u \in \{0, 1\}^{|\mathcal{X}|}$ over the item catalog $\mathcal{X}$. The architecture is asymmetric: the encoder is a deep, learnable network, while the decoder is a fixed, non-learnable content-based embeddings matrix $\bm{E} \in \mathbb{R}^{|\mathcal{X}| \times D}$:
\begin{itemize}
  \item \textbf{Encoder}: Maps the sparse history through multiple non-linear projections. The first layer is initialized with CLIP-based content embeddings to accelerate convergence and improve stability, but remains learnable to capture behavioral patterns. To prevent the model from simply memorizing the input and to force the extraction of latent features, we utilize wide intermediate layers ($4D$ and $2D$, where $D=512$) with ReLU activations. We apply $L_2$ regularization to all encoder weights, including the final projection into the latent bottleneck $\bm{z}_u$.
  \item \textbf{Fixed decoder}: To preserve high-inertia properties, the decoder is a frozen content-embedding matrix $\bm{E}$, ensuring that long-term embeddings remain constrained to the same fixed semantic basis introduced earlier in the LTE framework. There are no learnable parameters between the latent bottleneck $\bm{z}_u$ and the output layer.
\end{itemize}

The latent bottleneck $\bm{z}_u$ serves as the fine-tuned LTE. The reconstruction logits $\bm{\hat{y}}_u$ are computed via matrix multiplication with the fixed semantic basis:
\begin{equation*}
  \bm{\hat{y}}_u = \bm{z}_u \bm{E}^{\intercal}.
\end{equation*}
Training minimizes the binary cross-entropy reconstruction loss:
\begin{equation*}
  \mathcal{L} = - \frac{1}{|\mathcal{X}|} \sum_{j \in \mathcal{X}} \left[ h_{u,j} \log \sigma\left(\hat{y}_{u,j}\right) + \left(1 - h_{u,j}\right) \log\left(1 - \sigma(\hat{y}_{u,j})\right) \right],
\end{equation*}
where $\sigma(\cdot)$ is the sigmoid function. To mine ``harder'' negatives, we sample non-interacted items following a log-uniform popularity distribution, tuning the power factor to optimize AUC, recall, and precision.

\begin{figure}[t]
  \centering
  \includegraphics[width=1.0\linewidth]{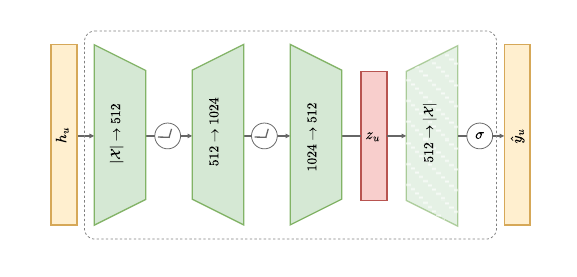}
  \Description{Architecture diagram of the asymmetric autoencoder. A learnable encoder maps a user's sparse multi-hot history vector through multiple non-linear layers into a latent bottleneck z_u. A frozen content-embedding matrix E serves as the decoder, producing reconstruction logits via dot product.}
  \caption{\textbf{Asymmetric autoencoder for LTE fine-tuning.} The learnable encoder projects a user's multi-hot history $\bm{h}_u$ into a latent bottleneck $\bm{z}_u$ (the fine-tuned LTE). By using a frozen content-embedding matrix $\bm{E}$ as the decoder, we force the model to learn behavioral affinities (e.g., price-point) while remaining grounded in a stable semantic space. Reconstruction is computed as $\bm{\hat{y}}_u = \bm{z}_u \bm{E}^{\intercal}$, ensuring compatibility with the heuristic content-based average.}
  \label{fig:asymmetric_autoencoder}
\end{figure}

For customers lacking behavioral data, the encoder is fed a zeroed multi-hot vector. In this scenario, the weight matrix of the first layer has no input features to project, leaving only the layer's bias vector to be processed. This bias acts as a learned default, representing the average customer profile or global popularity trends. As this signal propagates through the remaining encoder layers, it produces a constant LTE $\bm{z}_u$, which serves as a pre-computed baseline for cold-start users. Figure~\ref{fig:asymmetric_autoencoder} illustrates the architecture.

\subsection{Discussion and performance}

This design offers several systemic benefits for production environments. First, the fixed decoder constrains the latent space, ensuring cross-version compatibility and a seamless fallback to the content-based average. By forcing $\bm{z}_u$ to remain semantically aligned with the heuristic average, users with items outside the autoencoder's vocabulary are still represented within a familiar manifold. Second, the encoder learns collaborative behavioral patterns (behavioral weights), while the fixed decoder grounds outputs in contextual metadata (item attributes), thus combining the strengths of both collaborative and content-based approaches. Finally, since only the encoder is trainable, the model is significantly lighter than sequence-aware transformers, facilitating scaling across millions of users and high-cardinality catalogs.

Offline evaluations show that this behavioral fine-tuning yields a +2.1\% relative uplift in NDCG@500 over the uniform average baseline. This demonstrates that the autoencoder successfully identifies which interactions are most representative of a user's long-term taste without drifting from the stable content-based manifold.

In summary, this approach:
\begin{itemize}
  \item Captures customer affinities for style, product attributes, and price range from one year of behavioral data without being dominated by short-term intent.
  \item Maintains high-inertia properties by constraining the latent space to a fixed content-based semantic basis.
  \item Bridges collaborative and contextual expressiveness via the encoder-decoder split.
  \item Scales efficiently in production, as all trainable parameters are concentrated in the encoder.
  \item Provides a safe, calibrated default for new users through a learned bias-driven latent vector.
  \item Ensures fallback to a simple average for out-of-vocabulary users, as both methods reside in the same semantic space.
\end{itemize}

Future work includes online A/B testing of this fine-tuned LTE approach and a detailed comparison to heavier transformer-based methods for long-term embedding learning.

\section{Conclusion}

In this work, we addressed the fundamental tension between fluid short-term intent and stable long-term preference in sequential recommendation. While Transformer architectures excel at the former, their practical application for long-range history is limited by computational costs and a pronounced recency bias. We introduced a \textit{high-inertia LTE framework} designed specifically for the constraints of industrial production, where single-versioned feature stores necessitate temporal consistency and rollback resilience. Our findings demonstrate that grounding long-term signals in a \textit{fixed semantic basis} provides a robust contextual anchor for the model. Through contextual anchoring via a prefix token, we showed that the transformer effectively redistributes its attention budget, with patterns consistent with reclaiming distant memories and tempering short-term session volatility. Furthermore, we demonstrated that an \textit{asymmetric autoencoder} can fine-tune these representations on behavioral data without sacrificing the stability required for seamless model deployments. The effectiveness of our approach is validated by significant uplifts in both engagement (+0.61\%) and revenue (+0.42\%) in large-scale online A/B tests. By bridging the gap between short-term intent and long-term preference through a high-inertia framework, we provide a scalable, production-ready solution for more balanced and resilient personalization.

\begin{acks}
  We are grateful for the valuable feedback, insightful discussions, and constant support from our colleagues, as well as their contributions to the design and execution of the online experiments, including: Alisa Mironenko, Apolo Takeshi, Darya Dedik, Gabriel Coelho, Gayatri Kapur, Géraud Le Falher, Gokmen Oz, Hani Ahmad, Isa-Sertan Karabiyikli, Jacek Wasilewski, Jean-Baptiste Faddoul, Karthik Bappudi, Maarten Versteegh, Matti Lyra, Roberto Roverso, Satyajit Gupte, Stephen Redmond, and Ton Torres.
\end{acks}

\balance

\bibliographystyle{ACM-Reference-Format}
\bibliography{main}

\end{document}